\documentclass{article}

\usepackage[preprint,nonatbib]{neurips_2022}

\usepackage[utf8]{inputenc} 
\usepackage[T1]{fontenc}    
\usepackage{hyperref}       
\usepackage{url}            
\usepackage{booktabs}       
\usepackage{amsfonts}       
\usepackage{nicefrac}       
\usepackage{microtype}      
\usepackage{xcolor}   
\usepackage{amsmath}

\usepackage{amsfonts}
\usepackage{mathtools}
\usepackage{bm}
\usepackage{xfrac}
\usepackage{xspace}
\usepackage[makeroom]{cancel}
\usepackage{makecell}
\usepackage{multirow}
\newcommand*{\method}{DeVRF}
\usepackage{subcaption}

\usepackage{color}
\usepackage{xcolor}
\usepackage{colortbl}
\usepackage{lipsum}
\usepackage{ wasysym }
\usepackage{textcomp}

\usepackage[abs]{overpic}

\usepackage{booktabs}
\definecolor{myyellow}{rgb}{1,1, 0.6}
\definecolor{myorange}{rgb}{1, 0.8, 0.6}
\definecolor{myred}{rgb}{1, 0.6, 0.6}
\definecolor{blue-violet}{rgb}{0.54, 0.17, 0.89}

\title{DeVRF: Fast Deformable Voxel Radiance Fields for Dynamic Scenes}

\author{%
  Jia-Wei Liu$\textsuperscript{\rm 1}$\thanks{Work is partially done during internship at ARC Lab, Tencent PCG.},  Yan-Pei Cao$\textsuperscript{\rm 2}$, Weijia Mao$\textsuperscript{\rm 1}$, Wenqiao Zhang$\textsuperscript{\rm 4}$, David Junhao Zhang$\textsuperscript{\rm 1}$,  \\ 
  \textbf{Jussi Keppo$\textsuperscript{\rm 5,6}$, Ying Shan$\textsuperscript{\rm 2}$, Xiaohu Qie$\textsuperscript{\rm 3}$, Mike Zheng Shou$\textsuperscript{\rm 1}$\thanks{Corresponding Author.}} \\ 
  \\
  $~\textsuperscript{\rm 1}$  Show Lab, National University of Singapore $~\textsuperscript{\rm 2}$  ARC Lab,$~\textsuperscript{\rm 3}$  Tencent PCG\\
  $~\textsuperscript{\rm 4}$  National University of Singapore 
  $~\textsuperscript{\rm 5}$  Business School, National University of Singapore \\
  $~\textsuperscript{\rm 6}$  Institute of Operations Research and Analytics, National University of Singapore \\  
}

\begin{document}

\maketitle

\begin{abstract}

Modeling dynamic scenes is important for many applications such as virtual reality and telepresence. Despite achieving unprecedented fidelity for novel view synthesis in dynamic scenes, existing methods based on Neural Radiance Fields (NeRF) suffer from slow convergence (\emph{i.e.}, model training time measured in \emph{days}). In this paper, we present DeVRF, a novel representation to accelerate learning dynamic radiance fields. The core of DeVRF is to model both the 3D canonical space and 4D deformation field of a dynamic, non-rigid scene with explicit and discrete voxel-based representations. However, it is quite challenging to train such a representation which has a large number of model parameters, often resulting in overfitting issues. To overcome this challenge, we devise a novel \emph{static $\rightarrow$ dynamic} learning paradigm together with a new data capture setup that is convenient to deploy in practice. This paradigm unlocks efficient learning of deformable radiance fields via utilizing the 3D volumetric canonical space learnt from multi-view static images to ease the learning of 4D voxel deformation field with only few-view dynamic sequences. To further improve the efficiency of our DeVRF and its synthesized novel view's quality, we conduct thorough explorations and identify a set of strategies. We evaluate DeVRF on both synthetic and real-world dynamic scenes with different types of deformation. Experiments demonstrate that DeVRF achieves two orders of magnitude speedup (\textbf{\underline{100× faster}}) with on-par high-fidelity results compared to the previous state-of-the-art approaches. The code and dataset will be released in \url{https://github.com/showlab/DeVRF}.

\end{abstract}

\section{Introduction}

Free-viewpoint photorealistic view synthesis techniques from a set of captured
images unleash new opportunities for immersive applications such as virtual reality, telepresence, and 3D animation production. Recent advances in this domain mainly focus on static scenes, e.g., Neural Radiance Fields (NeRF) \cite{mildenhall2020nerf}, which implicitly represent rigid static scenes using 5D (spatial locations $\left(x,\,y,\,z\right)$ and view directions $\left(\theta,\,\varphi\right)$) neural radiance fields. Although achieving unprecedented fidelity for novel view synthesis, NeRF were mainly exploited under static scenes. To unlock dynamic view synthesis, existing NeRF-based approaches either learn an additional MLP-based deformation field that maps coordinates in dynamic fields to NeRF-based canonical spaces \cite{pumarola2021d,park2021nerfies,park2021hypernerf,tretschk2021non} or model dynamic scenes as 4D spatio-temporal radiance fields with relatively large MLPs \cite{li2021neural,gao2021dynamic}.

Despite being promising, NeRF is notoriously known for suffering from lengthy optimization time. This issue becomes particularly prominent for non-rigid, dynamic scenes because the aforementioned NeRF-based approaches require extra computation for the deformation MLPs \cite{pumarola2021d,park2021nerfies,park2021hypernerf,tretschk2021non} or time-varying texture/density querying \cite{li2021neural,gao2021dynamic}, resulting in quite long training time (in ``days''). 

\begin{figure}
\begin{centering}
\includegraphics[width=1\linewidth]{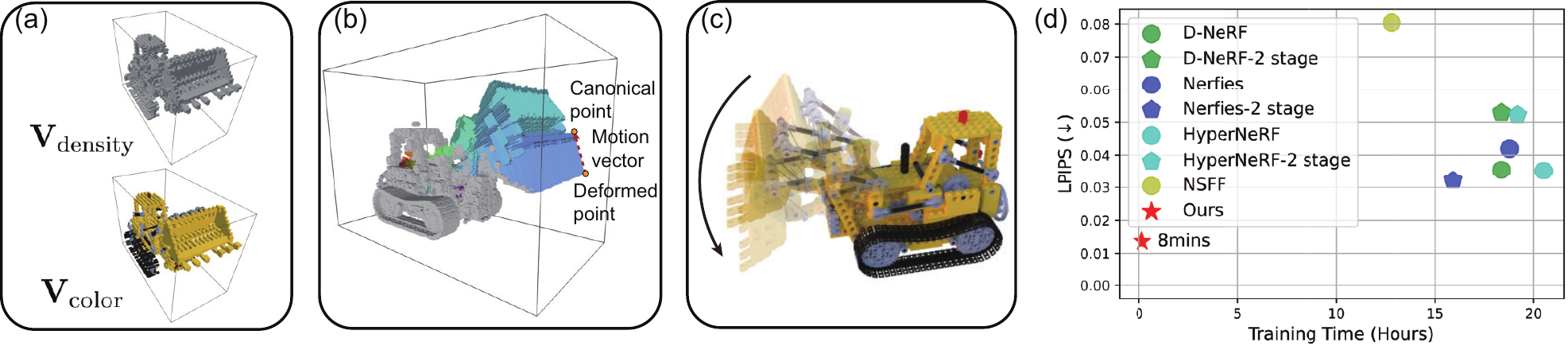}
\par\end{centering}
\caption{\label{fig:intro-model}The 3D canonical space \textbf{(a)} and the 4D deformation field \textbf{(b)} of DeVRF for neural modeling of a non-rigid scene \textbf{(c)}. \textbf{(d)}: The comparison between DeVRF and SOTA approaches. }
\end{figure}

This motivates us to improve the learning efficiency of dynamic radiance fields. Recent advances in static NeRF \cite{sun2021direct,yu2021plenoxels} show that employing voxel grids, such a volumetric representation, can achieve fast convergence. To adapt for dynamic scenes, one straightforward approach is to incorporate such a volumetric representation into the dynamic radiance field for fast neural modeling. In this paper, we present a novel deformable voxel radiance field (\method{}) that models both the 3D canonical space and 4D deformation field of a non-rigid, dynamic scene with explicit and discrete voxel-based representations, as illustrated in Fig. \ref{fig:intro-model} (a-c). However, we empirically observe that recklessly learning such a representation in dynamic radiance fields tends to plunge into the local optimum, \emph{i.e.}, the overfitting issue, due to the large number of parameters in DeVRF.

To overcome this overfitting issue, we power our DeVRF with two novel designs: \textbf{(1)} We devise an efficient and practical learning paradigm, \emph{i.e.}, \textbf{static $\rightarrow$ dynamic}, for learning deformable radiance fields. The key idea behind this is that the 3D volumetric canonical space learned from multi-view static images can introduce \emph{inductive bias}~\cite{baxter2000model} to unlock efficient learning of deformable radiance fields. 
Further, with such 3D priors, a dynamic scene can be effectively modeled with only a few fixed cameras. We argue that such a few-fixed-cameras setup for dynamic scene data capture is more convenient than the moving camera (such as the setup used in D-NeRF \cite{pumarola2021d}) in practice. 
\textbf{(2)} Based on the \emph{static $\rightarrow$ dynamic} paradigm,
we conduct extensive explorations and identify a set of strategies customised for DeVRF to improve its efficiency and effectiveness. These include a coarse-to-fine training strategy for the 4D deformation field to further improve efficiency, and three objectives to encourage our \method{} to reconstruct dynamic radiance fields with high fidelity: deformation cycle consistency, optical flow supervisions, and total variation regularization.

Fig. \ref{fig:intro-model} (d) shows that on five inward-facing synthetic scenes, two forward-facing real-world scenes and one inward-facing real-world scene, our approach enables fast dynamic radiance field modeling in about 10 minutes on a single NVIDIA GeForce RTX3090 GPU. This is \textbf{\underline{100$\times$}} faster than SOTA approaches with comparable novel view synthesis quality. 

To summarize, the major contributions of our paper are:




\vspace{-2mm}
\begin{itemize}
\item    
A novel perspective of DeVRF is presented that enables fast non-rigid neural scene reconstruction, which achieves an impressive 100$\times$ speedup compared to SOTA approaches with on-par high-fidelity.
\item
To the best of our knowledge, we are the first to incorporate the 4D voxel deformation field into dynamic radiance fields. 
\item
We devise a \emph{static $\rightarrow$ dynamic} learning paradigm that can boost performances with a low-cost yet effective capture setup.
\end{itemize}

\section{Related Work}


\textbf{Novel View Synthesis for Static Scenes.} Earlier approaches
\cite{kutulakos2000theory,seitz1999photorealistic,buehler2001unstructured,debevec1996modeling,10.1145/2980179.2982420,min2019vpmodel}
tackle novel view synthesis by first building an explicit 3D reconstruction 
of a static scene, such as voxels and meshes, and then rendering novel views
based on the reconstructed model. On the other hand, multi-plane images
\cite{zhou2018stereo,mildenhall2019local} represent a scene with multiple 
images at different depths and can reconstruct scenes with complex structures. 
Recently, NeRF \cite{mildenhall2020nerf}
achieves unprecedented fidelity for novel view synthesis by modeling static scenes with neural radiance fields. Subsequent
works have extended NeRF to different scenarios, such as few-view
novel view synthesis \cite{jain2021putting}, multi-scale representation \cite{barron2021mip}, and larger scenes \cite{tancik2022block,xiangli2021citynerf,rematas2021urban}.
However, these methods mainly focus on static scenes while the dynamic radiance fields reconstruction is more practical.

\textbf{Novel View Synthesis for Dynamic Scenes.} In order to capture dynamic scenes with non-rigidly deforming objects, 
traditional non-rigid reconstruction approaches require depth information as additional input or 
only reconstruct sparse geometry \cite{newcombe2015dynamicfusion,innmann2016volumedeform,yoon2020novel,collet2015high}. 
Neural Volumes \cite{lombardi2019neural} represents dynamic objects with a 3D voxel grid plus an implicit warp field, but requires an expensive multi-view capture rig and days to train.
Recent studies have built upon NeRF \cite{mildenhall2020nerf} and extended
it to dynamic neural radiance field reconstruction by learning
a mapping from dynamic to canonical field \cite{pumarola2021d,park2021nerfies,park2021hypernerf,tretschk2021non} or building a 4D spatio-temporal
radiance field \cite{xian2021space,li2021neural,gao2021dynamic,li2021neural3d}. D-NeRF \cite{pumarola2021d} learns
a deformation field that maps coordinates in a dynamic field to a
NeRF-based canonical space. Nerfies \cite{park2021nerfies} further
associates latent codes in the deformation MLP and the canonical NeRF
to tackle more challenging scenes such as moving humans. HyperNeRF \cite{park2021hypernerf}  proposes to
model the motion in a higher dimension space, representing the time-dependent
radiance field by slicing through the hyperspace.  In contrast, Video-NeRF \cite{xian2021space} models the dynamic scene as
4D spatio-temporal radiance fields and addresses motion ambiguity
using scene depth.
Sharing a similar idea on the 4D spatio-temporal field, NSFF \cite{li2021neural}
represents a dynamic scene as a time-variant function of geometry
and appearance, and warps dynamic scene with 3D scene motion. 
Lastly, several NeRF-based approaches have been proposed for 
modeling dynamic humans \cite{grassal2021neural,weng2022humannerf,liu2021neural,peng2021animatable,shao2022doublefield} but can not directly generalize to other scenes. 
Although achieving promising
results, existing methods require days of GPU training time, which is undesirable in real-world applications.

\textbf{NeRF Acceleration.} In the light of NeRF\textquoteright s
substantial computational requirements for training and rendering, recent papers have proposed methods 
to improve its efficiency. A line of work \cite{yu2021plenoctrees,reiser2021kilonerf,hedman2021baking} 
focuses on NeRF rendering acceleration and has achieved encouraging results. 
As for training acceleration, DVGO \cite{sun2021direct} models the radiance field with explicit 
and discretized volume representations, reducing training time to \emph{minutes}. Plenoxels \cite{yu2021plenoxels} employs 
sparse voxel grids as the scene representation and uses spherical harmonics to model view-dependent 
appearance, reaching a similar training speedup. Finally, Instant-ngp \cite{mueller2022instant} 
proposes multiresolution hash encoding; together with a highly optimized GPU implementation, 
it can produce competitive results after \emph{seconds} of training.
However, existing acceleration methods only focus on static scenes, 
while hardly any research, to our best knowledge, has studied NeRF acceleration 
for dynamic scenes. Very recently, Fourier PlenOctrees \cite{wang2022fourier} extends PlenOctrees \cite{yu2021plenoctrees} 
to dynamic scenes by processing time-varying density and color in the frequency domain; 
however, the data capturing setup is expensive, and it still requires hours of training. Instead, our proposed algorithm, 
\method{}, offers a superior training speed while only requires a few cameras for data capture.

\section{Method}

\subsection{Capture Setup}

Deformable scenes undergo various types of deformations and motions,
which can result in different scene properties such as object poses,
shapes, and occlusions. Therefore, capturing and modeling deformable
scenes is nontrivial even for professional photographic studios. Existing
approaches \cite{wang2022fourier,jiang20123d,zitnick2004high,bansal20204d}
attempt to capture 360{\textdegree} inward-facing dynamic
scenes with multi-view sequences and thus require dozens of high-quality
cameras. On the other hand, D-NeRF \cite{pumarola2021d} reconstructs
deformable radiance fields from a sparse set of synthetic images rendered
from a moving monocular camera. However, in practice, it is particularly
challenging to capture real-world 360{\textdegree} inward-facing dynamic scenes with a single moving camera due to various types
of deformations and resulting occlusions in dynamic scenes, especially
for scenes undergoing fast deformations. As a result, subsequent studies
\cite{park2021nerfies,park2021hypernerf,li2021neural,gao2021dynamic}
only capture forward-facing videos of real-world dynamic scenes with
a monocular camera.

Compared to dynamic scenes, it is much easier in practice to do multi-view 
capture for real-world static scenes with a monocular moving camera. 
Therefore, we propose to separate the capture process of a 
dynamic scene into two stages: the first stage captures a static state using 
a moving monocular camera, and the second stage captures the scene in motion 
using a few fixed cameras. In this capture setup,
the multi-view static images provide complete 3D geometry and appearance 
information of the scene, while few-view dynamic sequences show how the scene deforms
in 3D space over time; the entire capture process only requires a few cameras. 
Tab. \ref{tab:cameras} compares our capture process
with existing approaches in terms of the number of required cameras, cost,
and supported real-world use cases.

\begin{table}
\caption{\label{tab:cameras}Comparisons of capture setups for dynamic scenes.}

\resizebox{\linewidth}{!}{
\centering
\setlength{\tabcolsep}{1.8pt}
\begin{tabular}{cccc}
\hline 
Approach & No. of cameras & Cost & Supported real-world use cases\tabularnewline
\hline 
D-NeRF\cite{pumarola2021d}, Nerfies\cite{park2021nerfies} & Monocular & Low & Forward-facing scenes, slow reconstruction in \emph{days}.\tabularnewline
Neural Volumes\cite{lombardi2019neural} & Multiple (34) & High & 360{\textdegree} inward-facing scenes, slow reconstruction in \emph{days}.\tabularnewline
Fourier PlenOctrees\cite{wang2022fourier} & Multiple (60) & High & 360{\textdegree} inward-facing scenes, fast reconstruction
in $2\mathrm{hrs}$.\tabularnewline
\hline
\multirow{2}*{Ours} & \multirow{2}*{Few (4)} & \multirow{2}*{Low} & 360{\textdegree} inward-facing and forward-facing scenes,\tabularnewline
&&& super-fast reconstruction in $10\mathrm{mins}$.\tabularnewline

\hline 
\end{tabular}}
\end{table}

\subsection{Deformable Voxel Radiance Fields}

\begin{figure}
\begin{centering}
\includegraphics[width=1\linewidth]{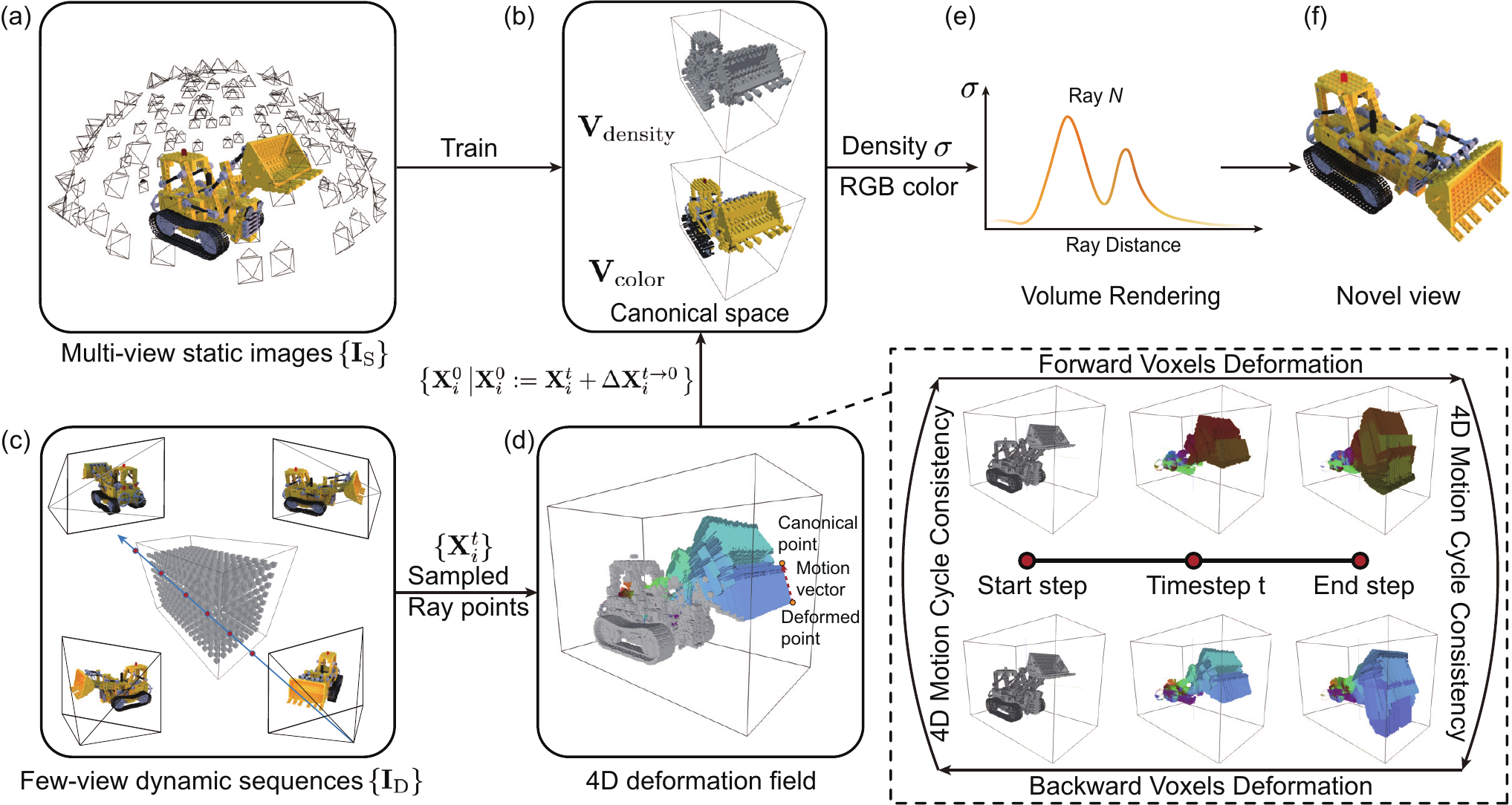}
\par\end{centering}
\caption{\label{fig:DeVRF}\textbf{Overview of our method.} In the first stage, DeVRF learns a 3D volumetric canonical prior \textbf{(b)} from multi-view static images \textbf{(a)}. In the second stage, a 4D deformation field \textbf{(d)} is jointly optimized from taking few-view dynamic sequences \textbf{(c)} and the 3D canonical prior \textbf{(b)}. For ray points sampled from a deformed frame, their deformation to canonical space can be efficiently queried from the 4D backward deformation field \textbf{(d)}. Therefore, the scene properties (\emph{i.e.}, density, color) of these deformed points can be obtained through linear interpolation in the 3D volumetric canonical space, and novel views \textbf{(f)} can be accordingly synthesized by volume rendering \textbf{(e)} using these deformed sample points.}
\end{figure}

As illustrated in Fig. \ref{fig:DeVRF}, we present DeVRF to model
both the 3D canonical space and 4D deformation field of a non-rigid
scene with explicit and discrete voxel-based representations. This
volumetric representation allows us to efficiently query the deformation,
density, and color of any 3D point at any time step in a deformable
scene, thus largely improving the training and rendering efficiency. 
In addition, we devise a \emph{static $\rightarrow$ dynamic}
learning paradigm that first learns a 3D volumetric canonical prior
from multi-view static images (Fig. \ref{fig:DeVRF}(a-b)) and transfers
such prior to dynamic radiance fields reconstruction (Fig. \ref{fig:DeVRF}(c-f)).


\textbf{3D Volumetric Canonical Space. }We take inspiration from the
volumetric representation of DVGO \cite{sun2021direct} and model
the scene properties such as density and color of our 3D canonical
space into voxel grids. Such representation enables us to efficiently
query the scene property of any 3D point via trilinear interpolation
of its neighboring voxels,
\begin{equation}
\mathrm{Tri\textrm{-}Interp}\left(\left[x,y,z\right],\,\mathbf{V}_{p}\right):\:\left(\mathbb{R}^{3},\,\mathbb{R}^{C\times N_{x}\times N_{y}\times N_{z}}\right)\rightarrow\mathbb{R}^{C}\,\label{eq:tri-interp}, \forall p \in \{\mathrm{density,\,color}\}
\end{equation}
\noindent where $C$ is the dimension of scene property $\mathbf{V}_{p}$. $N_x$, $N_y$ and $N_z$ are
the voxel resolutions of $\mathbf{V}_{p}$ in $x\textrm{-},\,y\textrm{-},\,z\textrm{-}$
dimension.

As shown in Fig. \ref{fig:DeVRF}(a-b), we learn the 3D volumetric
canonical prior, \emph{i.e.}, density grid $\mathbf{V}_{\mathrm{density}}$
and color grid $\mathbf{V}_{\mathrm{color}}$, with multi-view static
images $\left\{ \mathbf{I}_{\mathrm{S}}\right\} $ via volumetric
rendering. Following DVGO \cite{sun2021direct}, we employ $\mathrm{softplus}$
and post-activation after the trilinear interpolation of a 3D point
in $\mathbf{V}_{\mathrm{density}}$, as they are critical for sharp
boundary and high-frequency geometry reconstruction. We also apply
a shallow MLP after the trilinear interpolation of a 3D point in $\mathbf{V}_{\mathrm{color}}$
to enable view-dependent color effects \cite{sun2021direct}. In our
static $\rightarrow$ dynamic learning paradigm, the learned 3D volumetric
canonical prior provides critical knowledge of the 3D geometry and appearance
of the target dynamic scene, as few-view dynamic sequences alone struggle 
to reconstruct a complete deformable radiance field with high fidelity 
(as shown in Section \ref{sec:Experiments}).

\textbf{4D Voxel Deformation Field. }We employ a 4D voxel deformation field $\mathbf{V}_{\mathrm{motion}}$ to efficiently 
represent the motion of a deformable scene. As shown in Fig. \ref{fig:DeVRF}(d), the arrow directions represent the motions 
of voxels, the color denotes the motion direction, and the arrow magnitude denotes the motion scale. To synthesize a novel view at time step $t$, we shoot rays through image pixels and sample ray points $\mathcal{X}_t = \{\mathbf{X}_i^{t}\}$ in 3D space. The 3D motion  $\Delta\mathcal{X}_{t\rightarrow0}=\{\Delta\mathbf{X}_i^{t\rightarrow0}\}$ from $\mathcal{X}_t$ to their corresponding 3D points in the canonical space $\mathcal{X}_0 = \{\mathbf{X}_i^{0}~\rvert~\mathbf{X}_i^{0} = \mathbf{X}_i^{t} + \Delta\mathbf{X}_i^{t\rightarrow0}\}$ can be efficiently queried through quadruple interpolation of their neighboring voxels at neighboring time steps in the 4D backward deformation field,
\begin{equation}
\mathrm{Quad\textrm{-}Interp}\left(\left[x,y,z,t\right],\,\mathbf{V}_{\mathrm{motion}}\right):\:\left(\mathbb{R}^{4},\,\mathbb{R}^{N_{t}\times C\times N_{x}\times N_{y}\times N_{z}}\right)\rightarrow\mathbb{R}^{C}\,,
\end{equation}
\noindent where $C$ is the degrees of freedom (DoFs) of 
the sample point motion. We use $C=3$ in this paper, \emph{i.e.}, assign 
a displacement vector to each sample point.
$N_{t}$ is the number of key time steps that can be user-defined
based on the scene motion properties. 


Therefore, scene properties of $\mathcal{X}_t$ can then be obtained by 
querying the scene properties of their corresponding canonical points $\mathcal{X}_0$ 
through trilinear interpolation in the volumetric canonical space. Finally, pixel colors  can be calculated through volume rendering with the sampled scene properties along each ray, as illustrated in Fig. \ref{fig:DeVRF}(e-f).

\subsection{Optimization}

Training the DeVRF is quite challenging, mainly because a large number
of model parameters may lead to overfitting or suboptimal solutions.
This section describes the training strategy and optimization losses
that we design to facilitate fast optimization of the DeVRF.

\textbf{Coarse-to-Fine Optimization.} For a dense 4D voxel deformation
field with $N_{t}\times C \times N_{x}\times N_{y}\times N_{z}$ resolution,
there could be millions of free parameters, which are prone to overfitting
and suboptimal solutions. To solve this problem, we employ a coarse-to-fine
training strategy. Specifically, in our experiments, we progressively up-scale the $x\textrm{-}y\textrm{-}z$ resolution
of the 4D voxel deformation field from $10\times10\times10$ to $160\times160\times160$.
With this strategy, the 4D voxel deformation field first learns a
rough motion at the coarse stage, which is thereafter progressively refined
in finer stages. Our experiments demonstrate that the coarse-to-fine
strategy can effectively smoothen the optimization landscape of the 4D
voxel deformation field and remove most suboptimal solutions, 
thus largely improving the training efficiency and accuracy.

\textbf{Re-rendering Loss. }With sampled properties at $\mathcal{X}_t$, the color of a pixel can be calculated through volume rendering, \emph{i.e.}, by integrating the density and color of $\mathcal{X}_t$ along a ray $\mathbf{r}$ \cite{mildenhall2020nerf}:
\begin{equation}
\hat{C}\left(\mathbf{r}\right)=\sum_{i=1}^{N_r}T_{i}\left(1-\mathrm{exp}\left(-\sigma_{i}\delta_{i}\right)\right)c_{i}+T_{N_r+1}\mathbf{c}_{\mathrm{bg}},\:T_{i}=\mathrm{exp}\left(-\sum_{j=1}^{i-1}\sigma_{j}\delta_{j}\right)\,,\label{eq:render}
\end{equation}

\noindent where $N_r$ is the number of sampled deformed points along the ray,
$T_{i}$ represents the probability of light transmitting through
ray $\mathbf{r}$ to the $i\textrm{-}\mathrm{th}$ sampled point, and $1-\mathrm{exp}\left(-\sigma_{i}\delta_{i}\right)$
is the probability that light terminates at the $i\textrm{-}\mathrm{th}$
point. $\delta_{i}$ is the distance between adjacent sampled points, and
$\sigma_{i},\,c_{i}$ denote the density and color of deformed point $i$,
respectively. $\mathbf{c}_{\mathrm{bg}}$ is the pre-defined background
color.

Given the few-view training dynamic sequences with calibrated poses $\left\{ \mathbf{I}_{\mathrm{D}}\right\} $, DeVRF is optimized by minimizing the photometric MSE
loss between the observed pixels color $C\left(\mathbf{r}\right)$
and the rendered pixels color $\hat{C}\left(\mathbf{r}\right)$ :
\begin{equation}
\mathcal{L}_{\mathrm{Render}}=\frac{1}{\left|\mathcal{R}\right|}\sum_{\mathbf{r}\in\mathcal{R}}\left\Vert \hat{C}\left(\mathbf{r}\right)-C\left(\mathbf{r}\right)\right\Vert _{2}^{2}\,,
\end{equation}

\noindent where $\mathcal{R}$ is the set of rays in a mini-batch.

\textbf{4D Deformation Cycle Consistency.} As illustrated in Fig.
\ref{fig:DeVRF}(d), we enforce 4D deformation cycle consistency between backward and forward motion,
which regularizes the learned deformation field.
In the 4D deformation cycle, backward motion vectors $\Delta\mathcal{X}_{t\rightarrow0}$ models the motion from $\mathcal{X}_t$ to $\mathcal{X}_0$; in contrast,
forward motion vectors $\Delta\mathcal{X}_{0\rightarrow t}$
models the motion from $\mathcal{X}_0$ to their corresponding 3D points in the dynamic space $\tilde{\mathcal{X}}_t = \{\tilde{\mathbf{X}}_i^{t}~\rvert~ \tilde{\mathbf{X}}_i^{t} = \mathbf{X}_i^{0} + \Delta\mathbf{X}_i^{0\rightarrow t}\}$. The 4D motion cycle consistency 
can now be realized by minimizing the following cycle consistency loss $\mathcal{L}_{\mathrm{Cycle}}\left(t\right)$,
\begin{equation}
\mathcal{L}_{\mathrm{Cycle}}\left(t\right)=\frac{1}{2N_s}\sum_{i=1}^{N_s}\left\Vert \mathbf{X}^{t}_{i}-\tilde{\mathbf{X}}_i^{t}\right\Vert _{2}^{2}\,,\label{eq:cycle}
\end{equation}

\noindent where $N_s$ is the number of sampled 3D points in a mini-batch.

\textbf{Optical Flow Supervision.} The DeVRF is indirectly supervised by 2D optical flows estimated
from consecutive frames of each dynamic sequence using a pre-trained RAFT model \cite{teed2020raft}. For $\mathcal{X}_t$ and their corresponding $\mathcal{X}_0$, we first compute the corresponding 3D points of $\mathcal{X}_0$ at $t-1$ time step via forward motion $\tilde{\mathcal{X}}_{t-1} = \{\tilde{\mathbf{X}}_i^{t-1}~\rvert~ \tilde{\mathbf{X}}_i^{t-1} = \mathbf{X}_i^{0} + \Delta\mathbf{X}_i^{0\rightarrow t-1}\}$. After that, we project $\tilde{\mathcal{X}}_{t-1}$ onto the reference camera
and get their pixel locations $\tilde{\mathcal{P}}_{t-1}=\{\tilde{\mathbf{P}}_i^{t-1}\}$, and compute the induced optical flow with respect to the pixel location ${\mathcal{P}}_{t}=\{{\mathbf{P}}_i^{t}\}$
from which the rays of $\mathcal{X}_t$ are cast. We enforce the
induced flow to be the same as the estimated flow by minimizing $\mathcal{L}_{\mathrm{Flow}}\left(t\right)$,
\begin{equation}
\mathcal{L}_{\mathrm{Flow}}\left(t\right)=\frac{1}{\left|\mathcal{R}\right|}\sum_{\mathbf{r}\in\mathcal{R}}\sum_{i=1}^{N_r}w_{\mathbf{r},i}\left|\left(\tilde{\mathbf{P}}^{t-1}_{\mathbf{r},i}-\mathbf{P}^{t}_{\mathbf{r},i}\right)-\mathbf{f}_{\mathbf{P}^{t}_{\mathbf{r},i}}\right|\,,
\end{equation}

\noindent where 
$w_{\mathbf{r},i}=T_{i}\left(1-\mathrm{exp}\left(-\sigma_{i}\delta_{i}\right)\right)$
is the ray termination weights from Eq. (\ref{eq:render}),
and $\mathbf{f}_{\mathbf{P}^{t}_{\mathbf{r},i}}$ is the estimated 2D backward optical
flow at pixel $\mathbf{P}^{t}_{\mathbf{r},i}$.

\textbf{Total Variation Regularization.} We additionally employ a total variation
prior \cite{rudin1994total} when training the 4D voxel deformation field to
enforce the motion smoothness between neighboring voxels. At time step $t$,
\begin{equation}
\mathcal{L}_{\mathrm{TV}}\left(t\right)=\frac{1}{2\bar{N}}\sum_{i=1}^{\bar{N}}\underset{d\in C}{\sum}\left(\Delta_{x}^{2}\left(\mathbf{v}_{i}\left(t\right),\,d\right)+\Delta_{y}^{2}\left(\mathbf{v}_{i}\left(t\right),\,d\right)+\Delta_{z}^{2}\left(\mathbf{v}_{i}\left(t\right),\,d\right)\right)\,,
\end{equation}

\noindent where $\Delta_{x,y,z}^{2}$ is the squared difference of motion vectors between voxel $\mathbf{v}_{i}$ and its neighbors along $x, y, z$ axes. $\bar{N}=N_x \times N_y \times N_z$ denotes the number of voxels.

\textbf{Training Objective. }The overall training objective of DeVRF
is the combination of per-pixel re-rendering loss $\mathcal{L}_{\mathrm{Render}}$,
cycle consistency loss $\mathcal{L}_{\mathrm{Cycle}}$, optical flow loss
$\mathcal{L}_{\mathrm{Flow}}$, and total variation
regularization $\mathcal{L}_{\mathrm{TV}}$:
\begin{equation}
\mathcal{L}=\omega_{\mathrm{Render}}\cdot\mathcal{L}_{\mathrm{Render}}+\omega_{\mathrm{Cycle}}\cdot\mathcal{L}_{\mathrm{Cycle}}+\omega_{\mathrm{Flow}}\cdot\mathcal{L}_{\mathrm{Flow}}+\omega_{\mathrm{TV}}\mathcal{L}_{\mathrm{TV}}\,,
\end{equation}

where $\omega_{\mathrm{Render}},\,\omega_{\mathrm{Cycle}},\,\omega_{\mathrm{Flow}},\,\omega_{\mathrm{TV}}$
are weights for corresponding losses.

\section{Experiments\label{sec:Experiments}}

We extensively evaluate the DeVRF on various types of datasets, including
five synthetic\footnote{The \emph{Lego} scene is shared by NeRF \cite{mildenhall2020nerf}, licensed under the Creative Commons Attribution 3.0 License: https://creativecommons.org/licenses/by/3.0/. Other scenes are purchased from TurboSquid, licensed under the TurboSquid 3D Model License: https://blog.turbosquid.com/turbosquid-3d-model-license/.} 360{\textdegree} inward-facing dynamic
scenes, two real-world forward-facing dynamic scenes, and one real-world
360{\textdegree} inward-facing dynamic scene. We
run all experiments on a single NVIDIA GeForce RTX3090 GPU. During training, we set $\omega_{\mathrm{Render}}=1,\,\omega_{\mathrm{Cycle}}=100,\,\omega_{\mathrm{Flow}}=0.005$, and $\omega_{\mathrm{TV}}=1$
for all scenes.

\subsection{Comparisons with SOTA Approaches}
To demonstrate the performance of DeVRF, we compare DeVRF to various
types of SOTA approaches,  including a volumetric method Neural Volumes
\cite{lombardi2019neural}, NeRF-based methods D-NeRF \cite{pumarola2021d},
Nerfies \cite{park2021nerfies}, HyperNeRF \cite{park2021hypernerf},
and a time-modulated method NSFF \cite{li2021neural}. For a fair comparison,
since DeVRF follows a static $\rightarrow$ dynamic learning paradigm,
we additionally implement 2-stage versions of D-NeRF,
Nerfies, and HyperNeRF to learn a canonical
space prior in the first stage and then optimize a deformation network
in the second stage.
To show the effectiveness of our low-cost capture strategy for dynamic scenes, we also train these baselines using only a few-view dynamic sequences and observe a significant performance drop compared to those trained with both static and
dynamic data. For quantitative comparison,
peak signal-to-noise ratio (PSNR), structural similarity index (SSIM)
\cite{wang2004image}, and Learned Perceptual Image Patch Similarity
(LPIPS) \cite{zhang2018unreasonable} with VGG \cite{simonyan2014very} are employed
as evaluation metrics \footnote{Although improving test-time rendering speed is not the focus of our paper, DeVRF achieves $16\times\sim32\times$ test-time rendering speedup compared with other approaches, averaging 0.875 seconds per $540\times960$ image.}.

\newcommand{\tablefirst}[0]{\cellcolor{myred}}
\newcommand{\tablesecond}[0]{\cellcolor{myorange}}
\newcommand{\tablethird}[0]{\cellcolor{myyellow}}

\begin{table*}[t]
\centering
\caption{
Averaged quantitative evaluation on inward-facing synthetic and real-world scenes against baselines and ablations of our method. We color code each cell as \colorbox{myred}{\textbf{best}}, \colorbox{myorange}{\textbf{second best}}, and \colorbox{myyellow}{\textbf{third best}}. 
\label{tab:inwardfacing}
}
\resizebox{0.85\linewidth}{!}{
\centering
\setlength{\tabcolsep}{1.8pt}

\begin{tabular}{l||ccccc||ccccc}

\toprule
& \multicolumn{ 5 }{c||}{
  \makecell{
  \textsc{\small Synthetic inward-facing}
  }
}
& \multicolumn{ 5 }{c}{
  \makecell{
  \textsc{\small Real-world inward-facing }
  }
}
\\

& \multicolumn{1}{c}{ \footnotesize PSNR$\uparrow$ }
& \multicolumn{1}{c}{ \footnotesize SSIM$\uparrow$ }
& \multicolumn{1}{c}{ \footnotesize LPIPS$\downarrow$ }
& \multicolumn{1}{c}{ \footnotesize GPU (GB)$\downarrow$ }
& \multicolumn{1}{c||}{ \footnotesize Time$\downarrow$ }
& \multicolumn{1}{c}{ \footnotesize PSNR$\uparrow$ }
& \multicolumn{1}{c}{ \footnotesize SSIM$\uparrow$ }
& \multicolumn{1}{c}{ \footnotesize LPIPS$\downarrow$ }
& \multicolumn{1}{c}{ \footnotesize GPU (GB)$\downarrow$ }
& \multicolumn{1}{c}{ \footnotesize Time$\downarrow$ }

\\
\hline

  Neural Volumes~\cite{lombardi2019neural}
  &$9.620$
  &$0.532$
  &$0.5520$&$19.4$&$22.4\mathrm{hrs}$
  
  &$17.29$
  &$0.608$
  &$0.3440$ &$19.2$ &$22.0\mathrm{hrs}$

  \\  D-NeRF~\cite{pumarola2021d}
  &$31.83$
  &$0.960$
  &$0.0355$&$10.0$&$18.4\mathrm{hrs}$

  &$29.15$
  &$0.946$
  &$0.0643$ &$12.4$ &$22.1\mathrm{hrs}$

  \\  D-NeRF~\cite{pumarola2021d}-2 stage
  &$28.29$
  &$0.945$
  &$0.0528$&$9.7$&$18.4\mathrm{hrs}$
  
  &$27.21$
  &$0.936$
  &$0.0706$ &$13.2$ &$22.2\mathrm{hrs}$

  \\    D-NeRF~\cite{pumarola2021d}-dynamic
  &$17.59$
  &$0.839$
  &$0.2058$&$9.8$&$21.9\mathrm{hrs}$

  &$21.74$
  &$0.911$
  &$0.0906$ &$13.5$ &$22.3\mathrm{hrs}$

  \\   Nerfies~\cite{park2021nerfies}
  &$33.09$
  &\tablesecond$0.989$
  &$0.0432$&$21.8$&$18.7\mathrm{hrs}$
  
  &\tablethird$29.58$
  &\tablefirst$0.980$
  & $0.0576$&$22.5$ &$19.1\mathrm{hrs}$

  \\   Nerfies~\cite{park2021nerfies}-2 stage
  &$32.37$
  &\tablefirst$0.991$
  &$0.0322$&$22.0$&$15.8\mathrm{hrs}$

  &$23.93$
  &$0.920$
  &$0.0878$ &$22.0$ &$19.7\mathrm{hrs}$

  \\   Nerfies~\cite{park2021nerfies}-dynamic
  &$19.45$
  &$0.794$
  &$0.1674$&$22.0$&$21.3\mathrm{hrs}$

  &$20.70$  
  &$0.910$
  &$0.1080$ &$22.0$ &$19.6\mathrm{hrs}$

    \\ HyperNeRF~\cite{park2021hypernerf}
  &\tablethird$33.73$
  &$0.965$
  &$0.0335$&$22.5$&$20.5\mathrm{hrs}$

  &$28.50$
  &$0.944$
  &$0.0692$ &$22.0$ &$20.5\mathrm{hrs}$

     \\ HyperNeRF~\cite{park2021hypernerf}-2 stage
  &$29.16$
  &$0.953$
  &$0.0555$&$22.5$&$19.2\mathrm{hrs}$

  &$26.53$
  &$0.935$
  &$0.0802$ &$22.0$ &$19.3\mathrm{hrs}$

  \\ HyperNeRF~\cite{park2021hypernerf}-dynamic
  &$18.00$
  &$0.786$
  &$0.2173$&$22.4$&$20.6\mathrm{hrs}$

  &$10.39$
  &$0.734$
  &$0.3990$ &$22.0$&$20.5\mathrm{hrs}$

       \\ NSFF~\cite{li2021neural}
  &$27.06$
  &$0.936$
  &$0.0800$&$21.4$&$12.8\mathrm{hrs}$

  &$28.44$
  &$0.939$
  &$0.0714$ &$22.7$ &$15.3\mathrm{hrs}$

   \\ NSFF~\cite{li2021neural}-dynamic
    &$18.18$
  &$0.858$
  &$0.1929$&$15.0$&$15.5\mathrm{hrs}$

      &$19.90$
  &$0.909$
  & $0.0944$&$22.7$ &$16.2\mathrm{hrs}$

  \\ \hline  Ours (base)
  &$22.44$
  &$0.887$
  &$0.1173$&$4.6$&$8\mathrm{mins}$

  &$24.56$
  &$0.917$
  &$0.0844$&$6.5$&$10\mathrm{mins}$

  \\    Ours w/ c2f
  &$31.97$
  &$0.975$
  &$0.0185$&$4.6$&$7\mathrm{mins}$

  &$27.83$
  &$0.956$
  &$0.0465$&$6.5$&$10\mathrm{mins}$

  \\  Ours w/ c2f, tv
   &$32.73$
  &$0.963$
  &\tablethird$0.0172$&$4.6$&$7\mathrm{mins}$

  &$29.35$
  &$0.959$
  &\tablethird$0.0434$&$6.5$&$10\mathrm{mins}$

  \\   Ours w/ c2f, tv, cycle
  &\tablesecond$33.97$
  &$0.981$
  &\tablesecond$0.0142$&$4.6$&$8\mathrm{mins}$

  &\tablesecond$31.56$
  &\tablethird$0.971$
  &\tablesecond$0.0292$&$6.5$&$11\mathrm{mins}$

  \\   Ours w/ c2f, tv, cycle, flow
  &\tablefirst$34.29$
  &\tablethird$0.982$
  &\tablefirst$0.0137$&$4.6$&$8\mathrm{mins}$

  &\tablefirst$31.68$
  &\tablesecond$0.972$
  &\tablefirst$0.0289$&$6.5$&$11\mathrm{mins}$

  \\ \bottomrule

\end{tabular}

}
\vspace{-12pt}
\end{table*}

\begin{figure}
\begin{centering}
\includegraphics[width=1\linewidth]{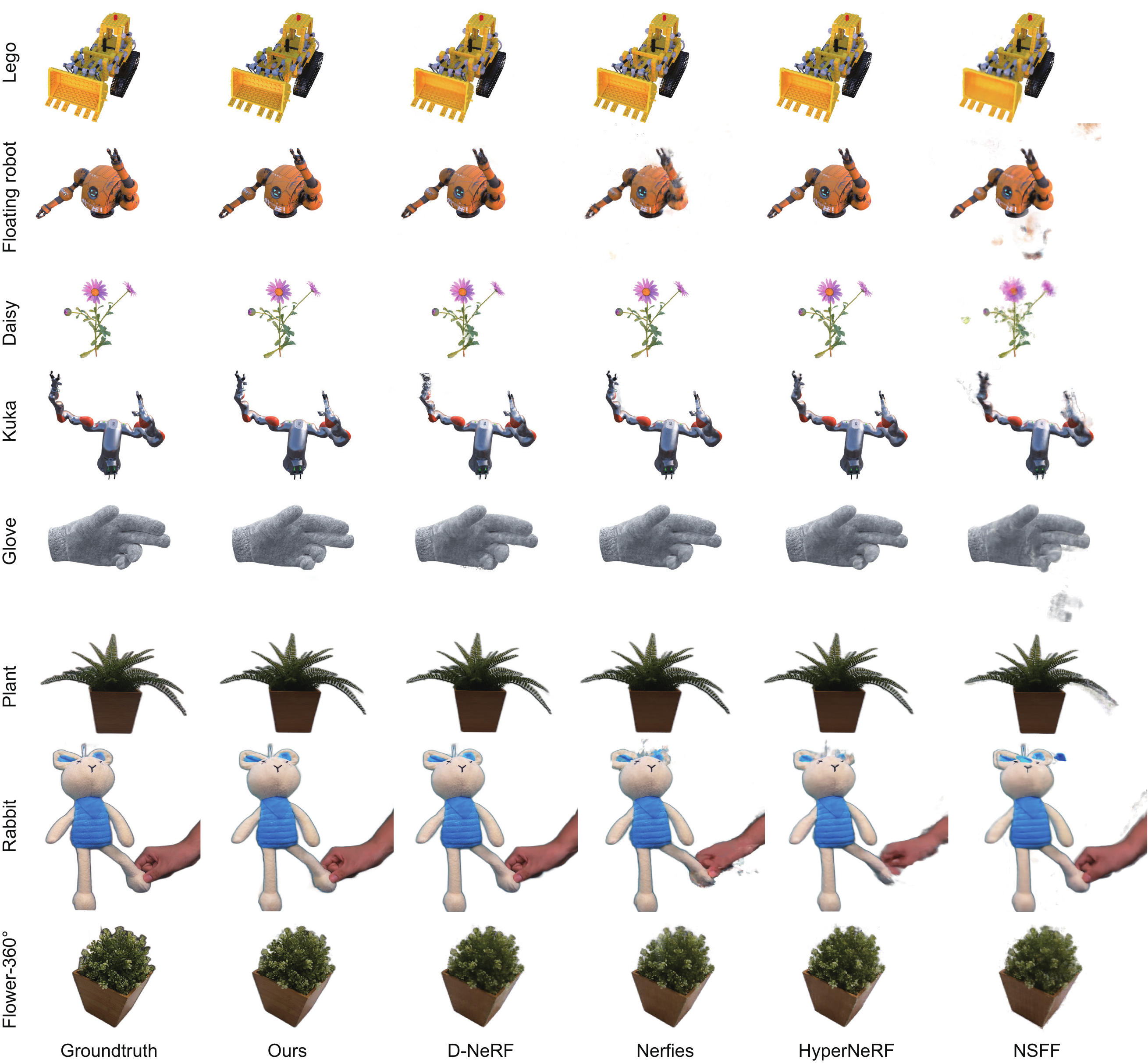}
\par\end{centering}
\caption{\label{fig:vis}Qualitative comparisons of baselines and DeVRF
on synthetic and real-world scenes.}
\end{figure}

\textbf{Evaluation on inward-facing synthetic and real-world deformable scenes.} We selected five synthetic dynamic scenes with various types of deformations and motions, and rendered synthetic images in $400\times400$ pixels under the 360{\textdegree} inward-facing setup.
For each scene, we use 100-view static images and 4-view  dynamic sequences with 50 frames (\emph{i.e.}, time steps) as training data for all approaches,
and randomly select another 2 views at each time step for test. In addition, we collected one 360{\textdegree} inward-facing real-world deformable scene in $540\times960$ pixels. With our data capture setup, only 4 cameras are required to capture dynamic scenes, and we choose 3 views of them as training data and the other view as test data.

We report the metrics of the real-world scene as well as the average metrics of five synthetic scenes for all approaches in Tab. \ref{tab:inwardfacing} and leave
the per-scene metrics to supplementary material. As shown in Tab. \ref{tab:inwardfacing}, for synthetic and real-world scenes, DeVRF achieves the best performance in terms of PSNR and LPIPS, and the second- or third-best in terms of SSIM among all approaches.
Most importantly, our per-scene optimization only takes less than $10\mathrm{mins}$ with $4.6\mathrm{GB}$  to $6.5\mathrm{GB}$  GPU memory on a single NVIDIA GeForce RTX3090 GPU, which is about two orders of magnitude faster than other approaches. The above quantitative comparison demonstrates the efficiency and effectiveness of DeVRF. Besides, the qualitative results of DeVRF and baselines on synthetic and real-world scenes are illustrated in Fig. \ref{fig:vis}, where DeVRF achieves on-par high-fidelity in comparison to SOTA methods. Please see the supplementary video for more results.

For a fair comparison, we additionally report the results of the 2-stage versions for D-NeRF
\cite{pumarola2021d}, Nerfies \cite{park2021nerfies}, and HyperNeRF
\cite{park2021hypernerf} in Tab. \ref{tab:inwardfacing}. Since these approaches are not designed to separately learn a canonical space and a deformation field, there is no significant difference in the results between their 2-stage versions and 1-stage versions using the training dataset. Furthermore, we also report the results of these baselines trained with dynamic data only (baseline-dynamic) in Tab. \ref{tab:inwardfacing}. Their performances drop significantly compared to the results trained with both static and dynamic data. In addition, since Neural Volumes \cite{lombardi2019neural} requires dozens of dynamic sequences as input, its performance is poor with our few-view dynamic sequences. The experimental results not only show the effectiveness of our low-cost data capture process and the proposed DeVRF model; but also validate our observation that few-view dynamic sequences alone fail to provide complete information about the dynamic scene, while static multi-view data can favorably serve as a supplement.


\begin{table*}[t]
\centering
\caption{
Quantitative evaluation on forward-facing real-world scenes against baselines and ablations of our system. We color code each cell as \colorbox{myred}{\textbf{best}}, \colorbox{myorange}{\textbf{second best}}, and \colorbox{myyellow}{\textbf{third best}}. 
\label{tab:forwardfacing}
}
\resizebox{0.85\linewidth}{!}{
\centering
\setlength{\tabcolsep}{1.8pt}

\begin{tabular}{l||ccccc||ccccc}

\toprule
& \multicolumn{ 5 }{c||}{
  \makecell{
  \textsc{\small Plant }
  }
}
& \multicolumn{ 5 }{c}{
  \makecell{
  \textsc{\small Rabbit }
  }
}

\\

& \multicolumn{1}{c}{ \footnotesize PSNR$\uparrow$ }
& \multicolumn{1}{c}{ \footnotesize SSIM$\uparrow$ }
& \multicolumn{1}{c}{ \footnotesize LPIPS$\downarrow$ }
& \multicolumn{1}{c}{ \footnotesize GPU (GB)$\downarrow$ }
& \multicolumn{1}{c||}{ \footnotesize Time$\downarrow$ }
& \multicolumn{1}{c}{ \footnotesize PSNR$\uparrow$ }
& \multicolumn{1}{c}{ \footnotesize SSIM$\uparrow$ }
& \multicolumn{1}{c}{ \footnotesize LPIPS$\downarrow$ }
& \multicolumn{1}{c}{ \footnotesize GPU (GB)$\downarrow$ }
& \multicolumn{1}{c}{ \footnotesize Time$\downarrow$ }

\\
\hline


  D-NeRF~\cite{pumarola2021d}
  &$31.94$
  &$0.979$
  &\tablethird$0.0251$&$11.4$&$21.5\mathrm{hrs}$
  
&\tablefirst$33.51$
  &\tablefirst$0.974$
&\tablefirst$0.0384$&$11.4$&$21.7\mathrm{hrs}$

  \\   Nerfies~\cite{park2021nerfies}
  &$31.36$
  &\tablefirst$0.991$
  &$0.0309$ &$21.9$ &$18.3\mathrm{hrs}$
  
  &$24.83$
  &$0.952$
  &$0.0795$ &$21.9$ &$19.8\mathrm{hrs}$

    \\ HyperNeRF~\cite{park2021hypernerf}
  &\tablefirst$32.08$
  &$0.978$
  &$0.0331$&$22.0$&$20.0\mathrm{hrs}$
  
  &$24.97$
  &$0.942$
  &$0.0849$&$22.0$&$20.7\mathrm{hrs}$

       \\ NSFF~\cite{li2021neural}
  &$29.45$
  &$0.966$
  &$0.0526$&$20.2$&$14.5\mathrm{\mathrm{hrs}}$
  
  &$27.68$
  &$0.945$
  &$0.0854$ & $20.2$&$14.5\mathrm{hrs}$

  \\ \hline  Ours (base)
  &$26.13$
  &$0.946$
  &$0.0722$&$8.3$&$8\mathrm{\mathrm{mins}}$
  
  &$25.58$
  &$0.910$
  &$0.1300$&$8.3$&$6\mathrm{mins}$

  \\    Ours w/ c2f
  &$31.85$
  &\tablethird$0.980$
  &$0.0275$&$8.3$&$8\mathrm{mins}$

  &$26.79$
  &$0.938$
  &$0.0946$&$8.3$&$6\mathrm{mins}$

  \\  Ours w/ c2f, tv
   &$31.89$
  &\tablethird$0.980$
  &$0.0263$&$8.3$&$8\mathrm{mins}$
  
  &$29.28$
  &$0.951$
  &$0.0655$&$8.3$&$6\mathrm{mins}$

  \\   Ours w/ c2f, tv, cycle
  &\tablethird$31.99$
  &\tablesecond$0.981$
  &\tablefirst$0.0235$&$8.3$&$9\mathrm{mins}$
  
  &\tablethird$31.05$
  &\tablethird$0.963$
  &\tablethird$0.0543$&$8.3$&$7\mathrm{mins}$

  \\   Ours w/ c2f, tv, cycle, flow
  &\tablesecond$32.01$
  &\tablesecond$0.981$
  &\tablesecond$0.0236$&$8.3$&$10\mathrm{mins}$
  
  &\tablesecond$32.05$
  &\tablesecond$0.966$
  &\tablesecond$0.0492$ &$8.3$ &$7\mathrm{mins}$

  \\ \bottomrule

\end{tabular}

}
\vspace{-12pt}
\end{table*}

\textbf{Evaluation on forward-facing real-world deformable scenes. }We collected
 two forward-facing real-world deformable scenes in $540\times960$ pixels using 4 cameras, and we chose 3 views of them as training data and the other view as test data. To handle forward-facing scenes, we adapt DeVRF to
use normalized device coordinates (NDC) and multi-plane images (MPI)
as in DVGO \cite{sun2021direct}. As shown in Tab. \ref{tab:forwardfacing}, DeVRF achieves the best result in the \emph{plant} scene and the second-best result in the \emph{rabbit} scene in terms of all metrics. Fig. \ref{fig:vis} also demonstrates qualitative comparisons on these two scenes.
\subsection{Ablation Study}

We carry out ablation studies on both synthetic and real-world scenes to
evaluate the effectiveness of each proposed component in DeVRF. We progressively
ablate each component from optical flow, cycle consistency, total
variation, to coarse-to-fine strategy. As shown in Tab. \ref{tab:inwardfacing} and \ref{tab:forwardfacing},
the performance of DeVRF progressively drops with the disabling of
each component, where disabling the coarse-to-fine training strategy
causes the most significant performance drop. This is as expected since the coarse-to-fine
training strategy is critical to reducing local minimums during optimization. 

\section{Conclusion}\label{sec:con}
We introduced DeVRF, a novel approach to tackle the challenging task of fast non-rigid radiance field reconstruction by modeling both the 3D canonical space and 4D deformation field of a dynamic scene with voxel-based representations. The DeVRF can be efficiently optimized in two major steps. We first proposed a \emph{static $\rightarrow$ dynamic} learning paradigm to pinpoint that the 3D volumetric canonical prior can be effectively transferred into the 4D voxel deformation field. Second, based on this learning paradigm, we developed a series of optimization strategies, including coarse-to-fine learning,  deformation cycle consistency, optical flow supervisions, and total variation priors. Such DeVRF finally produced a 100$\times$ faster training efficiency with on-par high-fidelity results in comparison to SOTA approaches. We believe our \method{} can provide a complement to existing literature and new insights into the view synthesis community.

\textbf{Limitations and Future Work. } Although DeVRF achieves fast deformable radiance field reconstruction, the model size is large due to its large number of parameters. In addition, DeVRF currently does not synchronously optimize the 3D canonical space prior during the second stage, and thus may not be able to model drastic deformations. We consider these limitations as faithful future work directions.

 \bibliographystyle{plain}
\bibliography{reference}

\appendix
\newpage

\section*{Appendix}
This supplementary material is organized as follows:

\begin{itemize}
\item   Section ~\ref{sec:1} provides more implementation details of the proposed \method{}.

\item Section ~\ref{sec:2} presents additional results of the per-scene evaluation.

\item Section ~\ref{sec:3} conducts additional ablations to further verify the effectiveness of \method{}.

\end{itemize}
In addition to this supplementary material, it is worth noting that we also provide a \textbf{supplementary video} to better visualize and compare our results to other SOTA approaches on all synthetic and real-world deformable scenes.

\section{Implementation Details}
\label{sec:1}
 We use the PyTorch~\cite{paszke2019pytorch} deep learning framework to conduct all our experiments on a single NVIDIA GeForce RTX3090 GPU.
 
\noindent\textbf{3D canonical space optimization.} During training, we set the voxel resolution of 3D canonical space, \emph{i.e.}, density grid $\mathbf{V}_{\mathrm{density}}$
and color grid $\mathbf{V}_{\mathrm{color}}$, to $160\times160\times160$ for inward-facing scenes and $256\times256\times128$ for forward-facing scenes, and we use a shallow MLP with $2$ hidden layers ($128$ channels for inward-facing scenes, and $64$ channels for forward-facing scenes). The 3D canonical space is optimized using a standard Adam optimizer~\cite{kingma2014adam} for $20\mathrm{k}$ with a batch size of $8192$ rays for inward-facing scenes and $4096$ rays for forward-facing scenes. The learning rate of $\mathbf{V}_{\mathrm{density}}$, $\mathbf{V}_{\mathrm{color}}$, and the designed MLP are set to $10^{-1}$, $10^{-1}$, and $10^{-3}$, respectively.

\noindent\textbf{4D voxel deformation field optimization.} The dense 4D voxel deformation field is modeled in $N_{t}\times C \times N_{x}\times N_{y}\times N_{z}$ resolution, which corresponds to $50\times 3 \times 160\times 160\times 160$. In our proposed coarse-to-fine optimization, we progressively upscale the $(x\textrm{-}y\textrm{-}z)$ resolution of the 4D voxel deformation field $\mathbf{V}_{\mathrm{motion}}$ as $(10\times10\times10)$ $\rightarrow$ $(20\times20\times20)$ $\rightarrow$ $(40\times40\times40)$ $\rightarrow$ $(80\times80\times80)$ $\rightarrow$ $(160\times160\times160)$. Such an optimization strategy can estimate the fine-grained voxels motion from a cascaded learning sequence. The base learning rate of the 4D voxel deformation field is $10^{-3}$, which is progressively decayed to $10^{-4}$ during coarse-to-fine optimization. For loss weights, we set $\omega_{\mathrm{Render}}=1,\,\omega_{\mathrm{Cycle}}=100,\,\omega_{\mathrm{Flow}}=0.005$, and $\omega_{\mathrm{TV}}=1$ across all scenes. The 4D voxel deformation field is optimized using Adam optimizer~\cite{kingma2014adam} for $25\mathrm{k}$ iterations with a batch size of $8192$ rays.

\section{Additional Results}
\label{sec:2}

\subsection{Per-scene Evaluation on Inward-facing Synthetic Deformable Scenes.} 

For quantitative comparison, Peak Signal-to-Noise Ratio (PSNR), Structural Similarity Index (SSIM)~\cite{wang2004image}, and Learned Perceptual Image Patch Similarity (LPIPS)~\cite{zhang2018unreasonable} with VGG~\cite{simonyan2014very} are employed
as evaluation metrics. PSNR and SSIM are simple and shallow functions, while LPIPS measures the perceptual similarity of deep visual representations and is more representative of visual quality. We report the per-scene comparisons on five inward-facing synthetic dynamic scenes - Lego, Floating robot, Daisy, Glove, Kuka - in Tab. \ref{tab:lego_robot_daisy} and Tab. \ref{tab:glove_kuka}. DeVRF achieves the best performance in terms of LPIPS in five scenes, and almost the second- or third-best in terms of PSNR and SSIM among all approaches. For the floating robot, daisy, and kuka, DeVRF achieves the best performance in terms of both the PSNR and LPIPS. Most importantly, our per-scene optimization only takes less than $10\mathrm{mins}$ with less than $5.0\mathrm{GB}$ GPU memory on a single NVIDIA GeForce RTX3090 GPU, which is about two orders of magnitude faster than other approaches.

\begin{table*}[ht]
\centering
\caption{
Per-scene quantitative evaluation on inward-facing synthetic scenes (Lego, Floating robot, and Daisy) against baselines and ablations of our method. We color code each cell as \colorbox{myred}{\textbf{best}}, \colorbox{myorange}{\textbf{second best}}, and \colorbox{myyellow}{\textbf{third best}}. 
\label{tab:lego_robot_daisy}
}
\resizebox{\linewidth}{!}{
\centering
\setlength{\tabcolsep}{1.8pt}

\begin{tabular}{l||ccccc||ccccc||ccccc}

\toprule
& \multicolumn{ 5 }{c||}{
  \makecell{
  \textsc{\small Lego }
  }
}
& \multicolumn{ 5 }{c||}{
  \makecell{
  \textsc{\small floating robot }
  }
}
& \multicolumn{ 5}{c}{
  \makecell{
  \textsc{\small daisy }
  }
}
\\

& \multicolumn{1}{c}{ \footnotesize PSNR$\uparrow$ }
& \multicolumn{1}{c}{ \footnotesize SSIM$\uparrow$ }
& \multicolumn{1}{c}{ \footnotesize LPIPS$\downarrow$ }
& \multicolumn{1}{c}{ \footnotesize GPU (GB)$\downarrow$ }
& \multicolumn{1}{c||}{ \footnotesize Time$\downarrow$ }
& \multicolumn{1}{c}{ \footnotesize PSNR$\uparrow$ }
& \multicolumn{1}{c}{ \footnotesize SSIM$\uparrow$ }
& \multicolumn{1}{c}{ \footnotesize LPIPS$\downarrow$ }
& \multicolumn{1}{c}{ \footnotesize GPU (GB)$\downarrow$ }
& \multicolumn{1}{c||}{ \footnotesize Time$\downarrow$ }
& \multicolumn{1}{c}{ \footnotesize PSNR$\uparrow$ }
& \multicolumn{1}{c}{ \footnotesize SSIM$\uparrow$ }
& \multicolumn{1}{c}{ \footnotesize LPIPS$\downarrow$ }
& \multicolumn{1}{c}{ \footnotesize GPU (GB)$\downarrow$ }
& \multicolumn{1}{c}{ \footnotesize Time$\downarrow$ }
\\
\hline

  Neural Volumes~\cite{lombardi2019neural}
  &$5.958$
  &$0.369$
  &$0.8314$&$19.4$&$22.4\mathrm{hrs}$
  
  &$6.403$
  &$0.405$
  &$0.7127$&$19.4$&$22.4\mathrm{hrs}$
  
  &$13.47$
  &$0.679$
  &$0.4429$&$19.4$&$22.4\mathrm{hrs}$

  \\  D-NeRF~\cite{pumarola2021d}
  &$28.41$
  &$0.935$
  &$0.0582$&$10.3$&$18.8\mathrm{hrs}$
  
  &$31.98$
  &$0.978$
  &$0.0251$&$10.3$&$18.4\mathrm{hrs}$
  
  &$33.51$
  &$0.990$
  &$0.0137$&$9.8$&$17.9\mathrm{hrs}$
  
  \\  D-NeRF~\cite{pumarola2021d}-2 stage
  &$24.34$
  &$0.885$
  &$0.1020$&$9.7$&$18.5\mathrm{hrs}$
  
  &$28.79$
  &$0.973$
  &$0.0289$&$9.7$&$18.5\mathrm{hrs}$
  
  &$31.40$
  &$0.985$
  &$0.0225$&$9.7$&$17.8\mathrm{hrs}$

  \\    D-NeRF~\cite{pumarola2021d}-dynamic
  &$20.29$
  &$0.852$
  &$0.1360$&$10.0$&$22.0\mathrm{hrs}$
  
  &$14.22$
  &$0.821$
  &$0.2720$&$10.0$&$21.6\mathrm{hrs}$
  
  &$22.76$
  &$0.947$
  &$0.0873$&$9.5$&$21.7\mathrm{hrs}$

  \\   Nerfies~\cite{park2021nerfies}
  &\tablesecond$30.34$
  &\tablefirst$0.986$
  &$0.0303$&$22.5$&$19.3\mathrm{hrs}$
  
  &$27.07$
  &$0.973$
  &$0.0773$&$22.5$&$18.4\mathrm{hrs}$
  
  &$38.26$
  &\tablefirst$0.998$
  &$0.0056$&$22.5$&$18.5\mathrm{hrs}$

  \\   Nerfies~\cite{park2021nerfies}-2 stage 
  &\tablethird$29.27$
  &\tablesecond$0.984$
  &$0.0449$&$22.4$&$15.8\mathrm{hrs}$
  
  &$30.05$
  &\tablefirst$0.991$
  &$0.0286$&$22.4$&$15.8\mathrm{hrs}$
  
  &$35.81$
  &\tablesecond$0.997$
  &$0.0100$&$22.4$&$15.8\mathrm{hrs}$

  \\   Nerfies~\cite{park2021nerfies}-dynamic
  &$21.15$
  &$0.911$
  &$0.1410$&$22.5$&$18.7\mathrm{hrs}$
  
  &$19.84$
  &$0.859$
  &$0.1400$&$22.5$&$19.3\mathrm{hrs}$
  
  &$23.71$
  &$0.864$
  &$0.1450$&$22.4$&$19.4\mathrm{hrs}$

    \\ HyperNeRF~\cite{park2021hypernerf}
  &\tablefirst$30.99$
  &$0.963$
  &$0.0360$&$22.5$&$21.3\mathrm{hrs}$
  
  &\tablethird$33.15$
  &$0.930$
  &$0.0511$&$22.5$&$19.8\mathrm{hrs}$
  
  &$36.31$
  &$0.994$
  &$0.0080$&$22.5$&$20.8\mathrm{hrs}$

     \\ HyperNeRF~\cite{park2021hypernerf}-2stage
  &$27.28$
  &$0.933$
  &$0.0758$&$22.5$&$21.3\mathrm{hrs}$
  
  &$28.28$
  &$0.955$
  &$0.0554$&$22.3$&$18.9\mathrm{hrs}$
  
  &$31.63$
  &$0.983$
  &$0.0238$&$22.3$&$18.5\mathrm{hrs}$

  \\ HyperNeRF~\cite{park2021hypernerf}-dynamic
  &$14.41$
  &$0.774$
  &$0.2910$&$22.4$&$19.8\mathrm{hrs}$
  
  &$14.88$
  &$0.835$
  &$0.2700$&$22.4$&$21.0\mathrm{hrs}$
  
  &$22.73$
  &$0.946$
  &$0.0957$&$22.4$&$21.0\mathrm{hrs}$

       \\ NSFF~\cite{li2021neural}
  &$25.44$
  &$0.912$
  &$0.0941$&$23.6$&$12.7\mathrm{hrs}$
  
  &$25.27$
  &$0.935$
  &$0.0944$&$20.9$&$13.2\mathrm{hrs}$

  &$28.71$
  &$0.967$
  &$0.0493$&$20.9$&$12.8\mathrm{hrs}$

   \\ NSFF~\cite{li2021neural}-dynamic
    &$15.14$
  &$0.762$
  &$0.2732$&$12.5$&$14.8\mathrm{hrs}$
  
    &$16.66$
  &$0.878$
  &$0.1769$&$15.7$&$16.0\mathrm{hrs}$
  
    &$24.02$
  &$0.937$
  &$0.1069$&$15.7$&$16.0\mathrm{hrs}$

  \\ \hline  Ours (base)
  &$17.83$
  &$0.799$
  &$0.2060$&$4.3$&$8\mathrm{mins}$
  
  &$21.74$
  &$0.907$
  &$0.1040$&$4.5$&$8\mathrm{mins}$
  
  &$25.91$
  &$0.950$
  &$0.0637$&$4.6$&$7\mathrm{mins}$

  \\    Ours w/ c2f
  &$27.55$
  &$0.952$
  &$0.0342$&$4.3$&$7\mathrm{mins}$
  
  &$32.04$
  &$0.983$
  &$0.0108$&$4.5$&$7\mathrm{mins}$
  
  &$37.89$
  &\tablethird$0.996$
  &$0.0055$&$4.6$&$6\mathrm{mins}$

  \\  Ours w/ c2f, tv
  &$28.44$
  &$0.958$
  &\tablethird$0.0301$&$4.3$&$8\mathrm{mins}$
  
  &$32.78$
  &$0.985$
  &\tablethird$0.0101$&$4.5$&$7\mathrm{mins}$
  
  &\tablesecond$38.55$
  &$0.950$
  &\tablethird$0.0048$&$4.6$&$6\mathrm{mins}$

  \\   Ours w/ c2f, tv, cycle
  &$29.11$
  &$0.963$
  &\tablesecond$0.0254$&$4.3$&$8\mathrm{mins}$
  
  &\tablesecond$34.12$
  &\tablethird$0.988$
  &\tablesecond$0.0084$&$4.5$&$8\mathrm{mins}$
  
  &\tablefirst$39.00$
  &\tablethird$0.996$
  &\tablefirst$0.0044$&$4.6$&$7\mathrm{mins}$

  \\   Ours w/ c2f, tv, cycle, flow
  &$29.25$
  &\tablethird$0.964$
  &\tablefirst$0.0250$&$4.3$&$9\mathrm{mins}$
  
  &\tablefirst$35.20$
  &\tablesecond$0.989$
  &\tablefirst$0.0074$&$4.5$&$9\mathrm{mins}$
  
  &\tablethird$38.39$
  &\tablethird$0.996$
  &\tablesecond$0.0046$&$4.6$&$7\mathrm{mins}$

  \\ \bottomrule

\end{tabular}

}
\end{table*}

\begin{table*}[h]
\centering
\caption{
Per-scene quantitative evaluation on inward-facing synthetic scenes (Glove and Kuka) against baselines and ablations of our method. We color code each cell as \colorbox{myred}{\textbf{best}}, \colorbox{myorange}{\textbf{second best}}, and \colorbox{myyellow}{\textbf{third best}}. 
\label{tab:glove_kuka}
}
\resizebox{0.75\linewidth}{!}{
\centering
\setlength{\tabcolsep}{1.8pt}

\begin{tabular}{l||ccccc||ccccc}

\toprule
& \multicolumn{ 5 }{c||}{
  \makecell{
  \textsc{\small glove }
  }
}
& \multicolumn{ 5 }{c}{
  \makecell{
  \textsc{\small kuka }
  }
}
\\

& \multicolumn{1}{c}{ \footnotesize PSNR$\uparrow$ }
& \multicolumn{1}{c}{ \footnotesize SSIM$\uparrow$ }
& \multicolumn{1}{c}{ \footnotesize LPIPS$\downarrow$ }
& \multicolumn{1}{c}{ \footnotesize GPU(GB)$\downarrow$ }
& \multicolumn{1}{c||}{ \footnotesize Time$\downarrow$ }
& \multicolumn{1}{c}{ \footnotesize PSNR$\uparrow$ }
& \multicolumn{1}{c}{ \footnotesize SSIM$\uparrow$ }
& \multicolumn{1}{c}{ \footnotesize LPIPS$\downarrow$ }
& \multicolumn{1}{c}{ \footnotesize GPU(GB)$\downarrow$ }
& \multicolumn{1}{c}{ \footnotesize Time$\downarrow$ }
\\
\hline

  Neural Volumes~\cite{lombardi2019neural}
  &$6.371$
  &$0.449$
  &$0.6101$&$19.4$&$22.4\mathrm{hrs}$
  
  &$15.92$
  &$0.757$
  &$0.1645$&$19.4$&$22.4\mathrm{hrs}$

  \\    D-NeRF~\cite{pumarola2021d}
  
  &$34.24$
  &$0.927$
  &$0.0455$&$9.8$&$18.5\mathrm{hrs}$

  &$31.03$
  &$0.975$
  &$0.0349$&$9.8$&$18.4\mathrm{hrs}$
  
  \\  D-NeRF~\cite{pumarola2021d}-2 stage

  &$30.86$
  &$0.922$
  &$0.0494$&$9.7$&$18.3\mathrm{hrs}$

  &$26.05$
  &$0.959$
  &$0.0614$&$9.6$&$18.8\mathrm{hrs}$

  \\    D-NeRF~\cite{pumarola2021d}-dynamic

  &$15.71$
  &$0.801$
  &$0.2660$&$10.3$&$22.1\mathrm{hrs}$

  &$14.97$
  &$0.777$
  &$0.2679$&$9.5$&$22.4\mathrm{hrs}$
  
  \\   Nerfies~\cite{park2021nerfies}
  
  &\tablefirst$36.37$
  &\tablefirst$0.993$
  &$0.0328$&$18.9$&$18.9\mathrm{hrs}$
  
  &\tablethird$33.40$
  &\tablefirst$0.996$
  &\tablethird$0.0193$&$22.5$&$18.5\mathrm{hrs}$
  
  \\   Nerfies~\cite{park2021nerfies}-2 stage 
  
  &\tablethird$34.96$
  &\tablesecond$0.991$
  &$0.0549$&$22.4$&$15.9\mathrm{hrs}$
  
  &$31.79$
  &\tablesecond$0.994$
  &$0.0227$&$22.4$&$15.8\mathrm{hrs}$
  
  \\   Nerfies~\cite{park2021nerfies}-dynamic 
  
  &$15.67$
  &$0.636$
  &$0.1740$&$22.5$&$18.5\mathrm{hrs}$
  
  &$16.86$
  &$0.698$
  &$0.2372$&$22.5$&$19.3\mathrm{hrs}$
  
    \\ HyperNeRF~\cite{park2021hypernerf}
  
  &\tablesecond$35.33$
  &$0.956$
  &$0.0471$&$22.5$&$20.0\mathrm{hrs}$
  
  &$32.88$
  &$0.983$
  &$0.0255$&$22.5$&$20.8\mathrm{hrs}$
  
     \\ HyperNeRF~\cite{park2021hypernerf}-2stage

  &$31.47$
  &$0.935$
  &$0.0694$&$22.3$&$18.8\mathrm{hrs}$
  
  &$27.12$
  &$0.959$
  &$0.0532$&$22.3$&$18.6\mathrm{hrs}$
  
  \\ HyperNeRF~\cite{park2021hypernerf}-dynamic

  &$22.04$
  &$0.850$
  &$0.1870$&$22.4$&$20.6\mathrm{hrs}$
  
  &$15.97$
  &$0.856$
  &$0.2429$&$22.4$&$20.8\mathrm{hrs}$

       \\ NSFF~\cite{li2021neural}

  &$27.66$
  &$0.902$
  &$0.1050$&$20.7$&$12.7\mathrm{hrs}$
  
  &$28.24$
  &$0.962$
  &$0.0574$&$20.9$&$12.7\mathrm{hrs}$
  
   \\ NSFF~\cite{li2021neural}-dynamic
  
  &$16.51$
  &$0.846$
  &$0.2091$&$15.7$&$16.0\mathrm{hrs}$

  &$18.59$
  &$0.865$
  &$0.1986$&$15.7$&$14.8\mathrm{hrs}$

  \\ \hline  Ours (base)

  &$25.14$
  &$0.867$
  &$0.1080$&$4.8$&$7\mathrm{mins}$

  &$21.59$
  &$0.914$
  &$0.1050$&$4.8$&$8\mathrm{mins}$
  
  \\    Ours w/ c2f
  
  &$30.69$
  &$0.959$
  &$0.0274$&$4.8$&$7\mathrm{mins}$

  &$31.70$
  &$0.985$
  &\tablefirst$0.0147$&$4.8$&$7\mathrm{mins}$
  
  \\  Ours w/ c2f, tv
  
  &$31.38$
  &$0.963$
  &\tablethird$0.0234$&$4.8$&$8\mathrm{mins}$
  
  &$32.48$
  &$0.987$
  &\tablesecond$0.0176$&$4.8$&$8\mathrm{mins}$
  
  \\   Ours w/ c2f, tv, cycle

  &$33.67$
  &$0.970$
  &\tablesecond$0.0183$&$4.8$&$8\mathrm{mins}$

  &\tablesecond$33.95$
  &\tablethird$0.989$
  &\tablefirst$0.0147$&$4.8$&$8\mathrm{mins}$
  
  \\   Ours w/ c2f, tv, cycle, flow
  
  &$34.67$
  &\tablethird$0.973$
  &\tablefirst$0.0168$&$4.8$&$8\mathrm{mins}$

  &\tablefirst$33.96$
  &\tablethird$0.989$
  &\tablefirst$0.0147$ &$4.8$&$9\mathrm{mins}$
  
  \\ \bottomrule

\end{tabular}

}
\end{table*}

\subsection{Per-scene Video Comparisons on Synthetic and Real-world Deformable Scenes.} 

We also provide a supplementary video to better visualize and compare our results to SOTA approaches on all five synthetic and three real-world deformable scenes. As can be seen from the video, our \method{} achieves on-par high-fidelity dynamic novel view synthesis results on all scenes and synthesizes the cleanest depth maps compared to other approaches. 

Significant quality enhancements of our \method{} can be observed in the video examples for floating robot, kuka, flower$\textrm{-}$360{\textdegree}, plant, and rabbit. Notably, clear differences can be observed in the plant and rabbit scenes, where D-NeRF~\cite{pumarola2021d} and NSFF~\cite{li2021neural} generate intermittent motions.
In contrast, the quadruple interpolation of the 4D voxel deformation field in our \method{} allows us to synthesize smooth motions at novel time steps.

In addition, the quadruple interpolation of the 4D voxel deformation field in \method{} allows us to conveniently and efficiently synthesize novel views at novel time steps, while existing approaches (i.e., Nerfies~\cite{park2021nerfies} and HyperNeRF~\cite{park2021hypernerf}) cannot synthesize the views at novel time steps that have not been seen during model training ~\cite{park2021nerfies,park2021hypernerf}. Thus, when rendering video examples, we generate results for Nerfies~\cite{park2021nerfies} and HyperNeRF~\cite{park2021hypernerf} only on the training and testing time steps. This makes their videos' duration shorter than ours.


\section{Additional Ablations}
\label{sec:3}

\begin{figure}[h]
\begin{centering}
\includegraphics[width=1.0\linewidth]{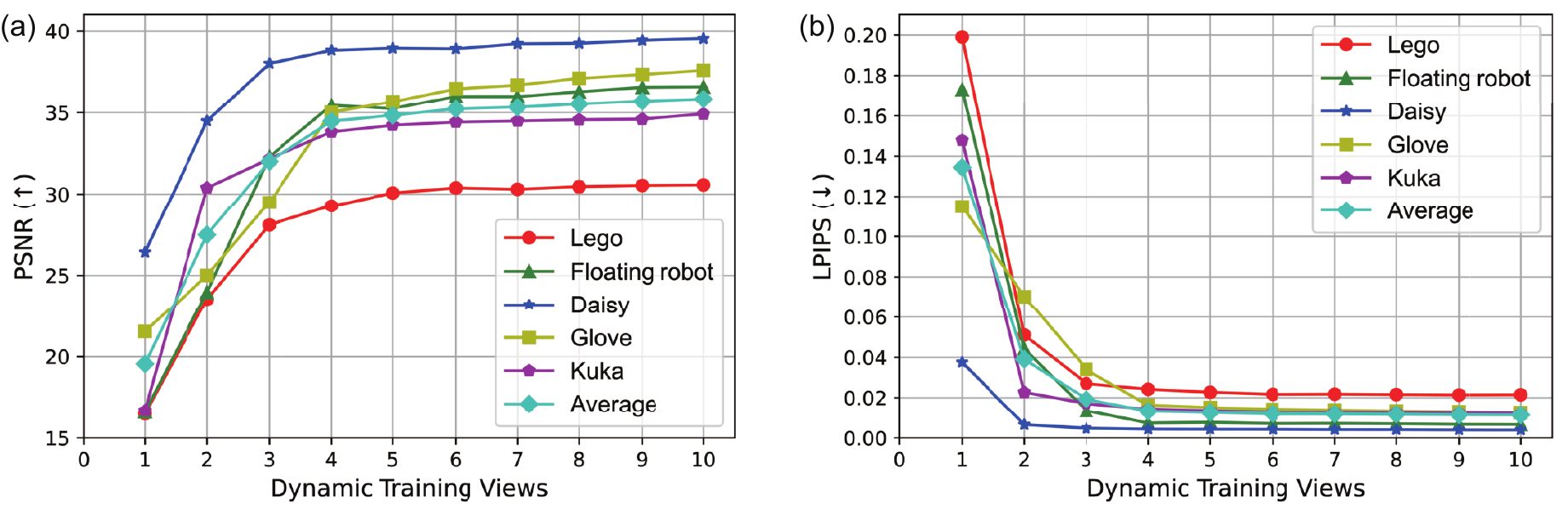}
\par\end{centering}
\caption{\label{fig:ablation_views}Ablation evaluation on the number of dynamic
training views: (a) PSNR, (b) LPIPS.}
\end{figure}

Our static $\rightarrow$ dynamic learning paradigm is based on a low-cost yet effective capture setup:  the multi-view static images provide complete 3D geometry and appearance information of the scene, while few-view dynamic sequences show how the scene deforms in 3D space over time. The entire capture process only requires a few cameras that are convenient to deploy in practice. To evaluate the influence of the number of dynamic views, we conduct additional ablations for DeVRF on five inward-facing synthetic dynamic scenes and report the per-scene metrics as well as the average metrics with respect to the number of dynamic training views. As shown in Fig. \ref{fig:ablation_views}, given the same multi-view static images, the performance of DeVRF largely improves with the increment of dynamic training views and almost saturates at six dynamic views, and the four dynamic training views used in our paper can yield comparable results compared to six dynamic views. Therefore, in our \emph{static $\rightarrow$ dynamic} learning paradigm, with static multi-view data as a supplement, only a few (e.g., four) dynamic views are required to significantly boost the performance of dynamic neural radiance fields reconstruction. This further demonstrates the effectiveness of our low-cost data capture process and the proposed DeVRF model.

\end{document}